\documentclass[conference]{IEEEtran}
\IEEEoverridecommandlockouts
\IEEEpubid{\makebox[\columnwidth]{978-1-6654-5674-6/22/\$31.00~\copyright2022 IEEE \hfill} \hspace{\columnsep}\makebox[\columnwidth]{ }}
\usepackage{cite}
\usepackage{amsmath,amssymb,amsfonts}
\usepackage{algorithm}
\usepackage{algorithmic}
\usepackage{graphicx}
\usepackage{textcomp}
\usepackage{xcolor}
\def\BibTeX{{\rm B\kern-.05em{\sc i\kern-.025em b}\kern-.08em
    T\kern-.1667em\lower.7ex\hbox{E}\kern-.125emX}}

\usepackage{subfigure}
\usepackage{colortbl}
\usepackage{ulem}
\usepackage{booktabs}
\usepackage{balance}

\usepackage{fancyhdr}
\usepackage{ragged2e}

\fancypagestyle{firstpage}
{
    \fancyhf{}
    \fancyhead[L]{\small \centering \noindent Preprint accepted at the 41st International Conference of the Chilean Computer Science Society, SCCC 2022, Santiago, Chile, 2022. 
    }
}

\begin{document}

\title{Reinforcement Learning for UAV control\\with Policy and Reward Shaping}

\author{\IEEEauthorblockN{Cristian Millán-Arias\IEEEauthorrefmark{1},
Ruben Contreras\IEEEauthorrefmark{2},
Francisco Cruz\IEEEauthorrefmark{3}\IEEEauthorrefmark{2}, and
Bruno Fernandes\IEEEauthorrefmark{1}}
\IEEEauthorblockA{\IEEEauthorrefmark{1}Escola Polit\'ecnica de Pernambuco, Universidade de Pernambuco, Recife, Brazil\\
Emails: \{ccma, bjtf\}@ecomp.poli.br}
\IEEEauthorblockA{\IEEEauthorrefmark{2}Escuela de Ingenier\'ia, Universidad Central de Chile, Santiago, Chile\\
Email: ruben.contreras@alumnos.ucentral.cl}
\IEEEauthorblockA{\IEEEauthorrefmark{3}School of Computer Science and Engineering, University of New South Wales, Sydney, Australia\\
Email: f.cruz@unsw.edu.au
}}

\maketitle
\thispagestyle{firstpage}
\IEEEpubidadjcol

\begin{abstract}
In recent years, unmanned aerial vehicle (UAV) related technology has expanded knowledge in the area, bringing to light new problems and challenges that require solutions. 
Furthermore, because the technology allows processes usually carried out by people to be automated, it is in great demand in industrial sectors.
The automation of these vehicles has been addressed in the literature, applying different machine learning strategies.
Reinforcement learning (RL) is an automation framework that is frequently used to train autonomous agents. 
RL is a machine learning paradigm wherein an agent interacts with an environment to solve a given task. 
However, learning autonomously can be time consuming, computationally expensive, and may not be practical in highly-complex scenarios. 
Interactive reinforcement learning allows an external trainer to provide advice to an agent while it is learning a task. 
In this study, we set out to teach an RL agent to control a drone using reward-shaping and policy-shaping techniques simultaneously. 
Two simulated scenarios were proposed for the training; one without obstacles and one with obstacles. We also studied the influence of each technique. 
The results show that an agent trained simultaneously with both techniques  obtains a lower reward than an agent trained using only a policy-based approach. 
Nevertheless, the agent achieves lower execution times and less dispersion during training.

\end{abstract}


\section{Introduction}
In recent years, machine learning has dominated the data science market, as it seeks to solve problems related to people’s daily lives, where machines learn specific tasks not through programmed instructions but by learning autonomously~\cite{bishop2006pattern}.
For many years, robotic technologies focused solely on industrial applications. 
However, the incredible technological growth in hardware and software capabilities has expanded the scope of robotics applications. %; one notable example can be seen in its applications in the agricultural industry~\cite{agri}. 
%As mentioned above, 

Technological progress goes hand in hand with the growth of knowledge in the area as well as the new challenges that emerge from time to time and, with these, new strategies to address them. 
One item that has sparked attention in recent years is the use of unmanned aerial vehicles (UAV), which are increasingly in demand in the market in general. 
UAVs were first created about a century ago, around 1916. While working for the British Air Ministry, Archibald Low dedicated himself to a project that aimed to develop a defense against German airships~\cite{mills2019dawn}. 
Today, in the 21st century, UAVs have taken on important roles in many areas, such as military applications, agriculture, recreation, etc.~\cite{drone1}. 
These practical devices can execute flyovers of areas while recording images which, in some cases, are transmitted in real-time.
UAVs operate autonomously, that is, they follow a trajectory or perform a task without the need for human intervention. 
Thus, one of the main challenges lies in the precision of the models used to control the vehicle and describe the target. 
Due to the lack of prior information or limited knowledge of the environment, the implementations that are achieved are unrealistic and not applicable in a real environment~\cite{contreras2020unmanned}.

Reinforcement learning (RL) is an approach that seeks to resolve the problem of an agent interacting with the environment to learn the desired task autonomously. 
The agent learns from its own experience, taking actions and discovering which of them yields the greatest reward~\cite{sutton2018reinforcement}. 
One problem with RL that has yet to be resolved is the time an RL agent spends learning~\cite{cruz2014improving}. 
Allowing an agent to find the ideal policy is excessively time consuming, mainly due to the large and complex state and action spaces. 
Furthermore, the agent can explore different areas of the space in order to find the space-action pair that yields the greatest reward~\cite{Thomaz2006a}.
To overcome this issue, an RL agent can be guided by a trainer in order to complete the task more quickly. 
Interactive reinforcement learning (IRL)~\cite{cruz2017agent, millan2019human} is an approach in which an external agent provides advice to the agent, whether to guide which action must be taken or to indicate if the correct decision was made.

In this study, we set out to teach a reinforcement learning agent to operate a drone in a simulated environment, complementing the learning with interactive reinforcement learning techniques known as reward-shaping and policy-shaping. 
The trainer provides advice using domain-based speech recognition~\cite{contreras2020unmanned}, where audio is selected from a set of recordings and processed in a phoneme matching algorithm. 
The trainer thus provides the reward or action in accordance with the approach applied, be it reward-shaping or policy-shaping, respectively. 
Two simulated scenarios were proposed for the training; one without obstacles and one with obstacles that impede the movement of the drone. 
The influence of the reward-shaping and policy-shaping approaches were analyzed individually and together, and were compared with an autonomous agent. 
A comparison was made of the average reward of the different agents as well as the execution time of each approach.
This paper is organized as follows: Section 2 presents an overview of reinforcement learning. 
Section 3 contains the methodology and proposed architecture of the study. 
Section 4 discusses the configuration of the experiments. 
Section 5 presents the main results. Lastly, Section 6 presents the conclusions and future studies to be developed.

\section{Reinforcement Learning and Interactive Feedback}

\begin{figure}[t]
  \centering
    \includegraphics[width=0.8\linewidth]{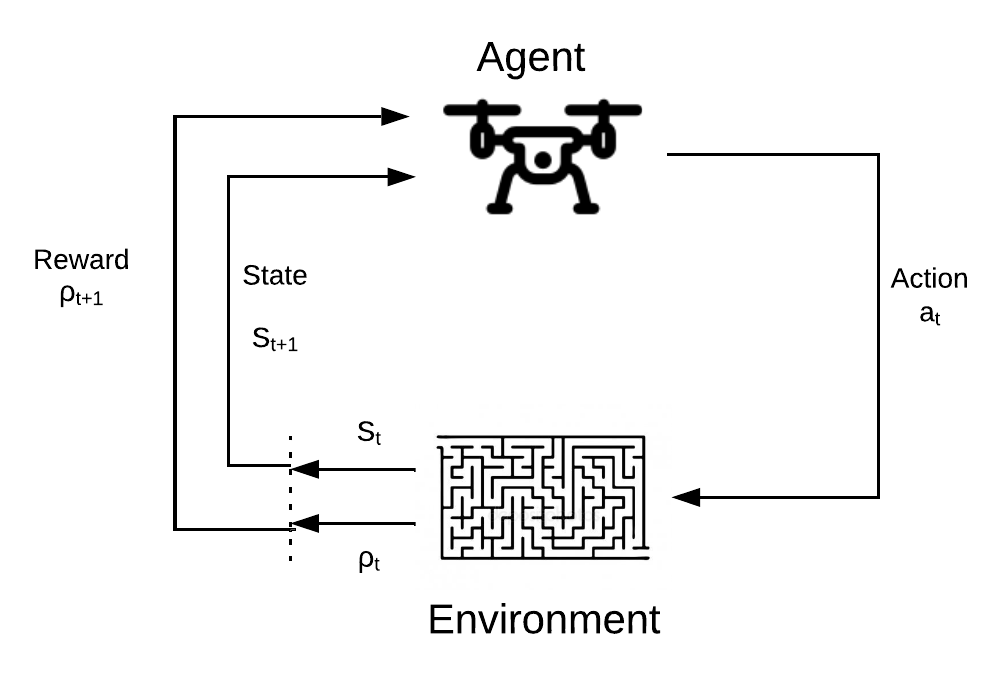}
   \caption{Interaction between the agent and the environment in a reinforcement learning context. The agent selects an action $a_{t}$ in state $s_{t}$, achieving a new state $s_{t+1}$ and a numeric reward $\rho_t$~\cite{sutton2018reinforcement}.}
   \label{Fig:RL}
\end{figure}

Reinforcement learning (RL) is an approach in the science of machine learning that allows new tasks to be learned through observing the reactions in an environment to actions taken. 
It is a behavioral learning model in which the algorithm learns through trial and error and is given feedback on its actions, allowing the agent to identify whether it is making good or bad decisions. 
This type of algorithm seeks to simulate natural learning and how it is put into practice in cognitive beings. 
In this approach, a learning agent interacts with its environment, as can be seen in Fig.~\ref{Fig:RL}. 
As it interacts, it generates knowledge, which helps it to make decisions going forward.

\subsection{Markov Decision Process}

Markov decision processes (MDP) provide a simple description of reinforcement learning tasks.
An MDP is defined by the tuple $< S, A_s, f, \rho >$ where~\cite{Knox2011}:

\begin{itemize}
\item $S$ is the set of states.
\item $A_s$ is the set of actions available in state $S$.
\item $f$ is the transition function, $f:S\times A \longrightarrow S$, which produces the new state given the selected action in the current state.
\item $\rho$ is the reward function, $\rho: S\times A\times S \rightarrow \mathbb{R}$, which evaluates the immediate performance of the selected action.
\end{itemize}

In an MDP, the transitions and rewards depend only on the current state and action selected by the agent~\cite{markov2014, cruz2018action}. 
A Markov state contains all the information related to the dynamics of a task. 
That is, once the current state is known, the history of the transitions that brought the agent to that position is irrelevant in terms of the decision-making problem.
The action $a_t$ taken in state $s_t$ is selected according to a policy $\pi$. The policy may be deterministic or stochastic. 
A deterministic policy is defined as a function that maps actions to states. 
A stochastic policy follows a distribution function, so actions are randomly selected from $\pi$ given the current state $s_t$.
Solving an MDP involves finding the policy that minimizes an optimality criterion, specifically, maximizing an expected return~\cite{Sigaud2010}.

\subsection{Temporal difference and Q-learning algorithm}

Temporal difference (TD) is a central part of RL methods. 
TD learning does not require a model of the environment to function since the agent learns with each action and not when an episode ends~\cite{sutton2018reinforcement}. 
TD-based algorithms attempt to generalize the relationship between the immediate reward and the prediction of a reward in the next iteration.
When the new iteration takes place, the reward obtained is compared to the reward expected. 
This difference, known as the temporal difference, is used to adjust the new prediction to the previous prediction~\cite{dabney2020distributional}.

The Q-learning algorithm~\cite{watkins1992q} is an off-policy method based in TD learning. In the Q-learning algorithm, the value function $Q(s_t, a_t)$ is updated by value $\max_{a\in A(s_{t+1})}Q(s_{t+1}, a_t)$.
The algorithm is based on the Bellman equation, an update rule that takes the weighted average of the previous value and the new information for the state:

\begin{multline}
Q(s_t, a_t) \leftarrow Q(s_t, a_t) + \\
\alpha \left[ \rho_{t+1} + \gamma \max_{a\in A(s_{t+1})}Q(s_{t+1}, a_t) - Q(s_t, a_t) \right],
\end{multline}

\noindent where $\alpha$ is the learning rate, and $\gamma$ the discount factor.
The value $Q(s_t,a_t)$ estimates the state-action pair once the action is applied in the current state. 
The learned value function, $Q$, directly approximates the optimal value function $Q^*$, independent of the policy~\cite{sutton2018reinforcement}.

\subsection{Interactive Reinforcement Learning}

In some situations, allowing an agent to learn a task by itself is impractical due to the high cost of each trial and error. 
Furthermore, learning autonomously can involve exploration problems and a slight tendency to avoid finding an optimal policy~\cite{Knox2009}. 
Interactive Reinforcement Learning (IRL) is an approach that considers the inclusion of a trainer with knowledge of the environment that guides the agent in order to optimize the learning time. 
The advice can be come from an expert or non-expert trainer, from artificial agents with full knowledge of the task, or from previously trained agents~\cite{cruz2014improving, bignold2021evaluation}. 
In an IRL scenario, it is expected that the interaction between the trainer and the agent will be as minimal as possible, otherwise, it would constitute supervised learning~\cite{nguyen2021broad, bignold2021persistent}.

%Figura RewardShaping
\begin{figure}[t]
  \centering
    \includegraphics[width=1.0\linewidth]{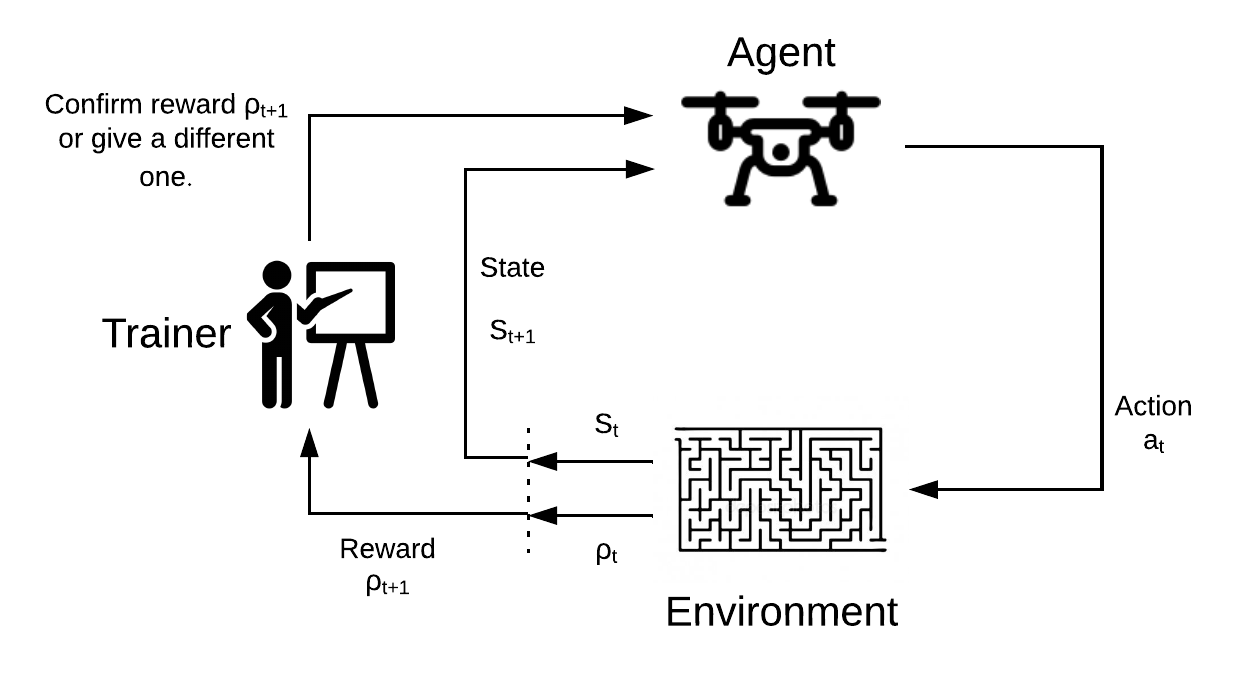}
   \caption{Reward-shaping: Interaction between the agent and the environment, wherein an external trainer intervenes in the reward received from the environment.}
   \label{Fig:Reward}
\end{figure}

\begin{figure}[t]
  \centering
    \includegraphics[width=1.0\linewidth]{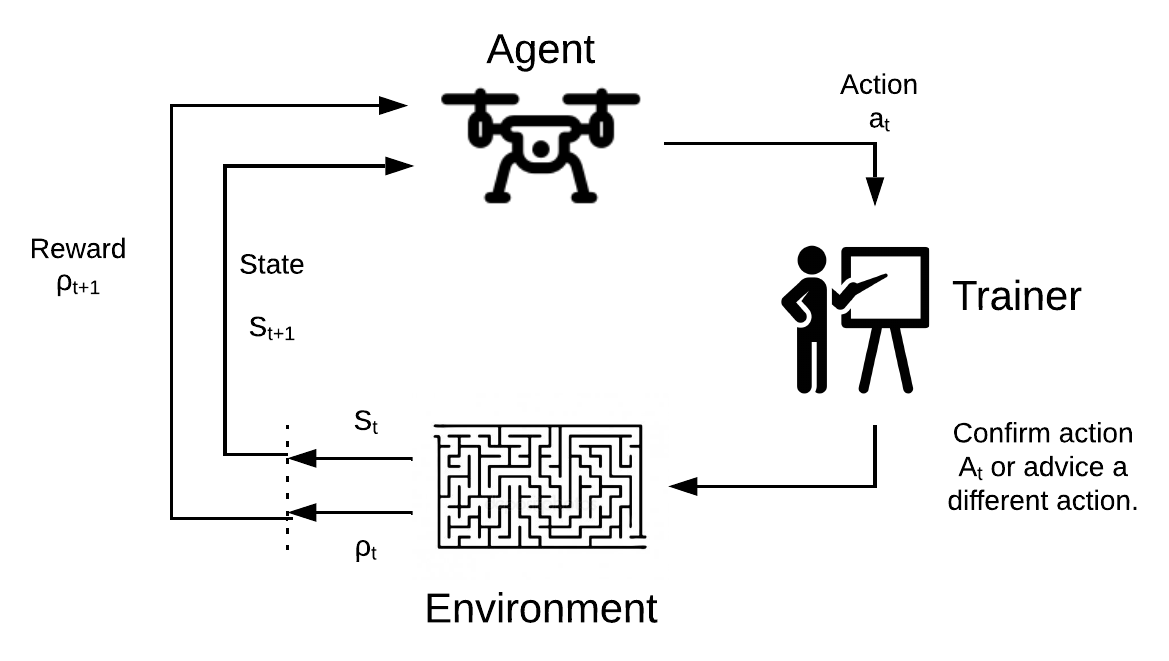}
   \caption{Policy-shaping: Interaction between the agent and the environment, wherein an external trainer intervenes in the agent's selected action. }
   \label{Fig:Policy}
\end{figure}

In terms of IRL, there are two distinct forms of relationship between the trainer and the agent.
In the first, known as reward-shaping, the trainer modifies the reward the agent receives from the environment. 
Fig.~\ref{Fig:Reward} illustrates the reward-shaping method, where the external trainer can modify the reward obtained from the environment. 
The reward from the trainer informs the agent about its performance for the selected action in the previous iteration~\cite{Thomaz2007,churamani2020icub}. 
In the second method, known as policy-shaping, the trainer modifies the action selected by the agent and suggests an action to be applied. 
Fig.~\ref{Fig:Policy} shows the policy-shaping method, where the external trainer proposes a new action to be performed, instead of the action selected by the agent. 
In this approach, it is expected that the action given by the trainer will perform better than that of the agent, thus increasing the probability it will be selected in the future~\cite{Thomaz2006a, Knox2009}.

It should be noted that reward-shaping modifies the estimate of the expected reward, giving a greater value to high-performance actions; however, it can be problematic for other types of actions. 
In policy-shaping, it is the policy of the agent that changes, rather than the reward function being affected~\cite{bignold2020conceptual}; however, the agent may take time to find an optimal policy due to the quality of advice received~\cite{Griffith13, bignold2021persistent}.

There are different papers that discuss reward-shaping and policy-shaping during the training of an agent. 
In terms of reward-shaping, Xiao el al.~\cite{xiao2020fresh} proposed a methodology called feedback-based reward-shaping (FRESH), which effectively integrates human feedback with deep RL algorithms in high-dimensional spaces. 
It uses a feedback neural network that effectively generalizes the feedback signals provided by the human trainer. 
Okudo et al.~\cite{okudo2021subgoal} proposed a subgoal-based reward shaping method that accelerates learning by using subgoal knowledge obtained from the original remaining optimal policy. 
Obtaining subgoals does not require manipulating the agent, and they can be obtained from failed MDPs. 
Passalis et al.~\cite{passalis2020continuous, tzimas2020leveraging} proposed a method for fine-grained UAV control in continuous spaces using deep RL for the purposes of acquiring high-quality frontal view person shots. 
The authors use a reward-shaping methodology to modify sparse rewards and improve the stability of the RL algorithm employed.

With respect to policy-shaping, Lin et al.~\cite{lin2017explore} propose a deep RL technique to learn from human advice that may be inconsistent or intermittent. 
They applied the methodology in 3D worlds, where the reward for the environment and the human feedback are on different scales. 
Harrison et al.~\cite{harrison2017guiding} propose a technique that uses natural language to help generalize environments that are unseen by the RL agent. 
The authors use encoder-decoder networks to learn associations between natural language descriptions and information from the environment. 
Navidi et al.~\cite{navidi2021new} combine reinforcement learning and imitation learning. 
In their study, they employ a cooperative structure where feedback from the trainer – part of imitation learning – is used to improve the behavior of the agent.
Lastly, Bignold et al.~\cite{bignold2022human} carry out a comparison between reward-shaping and policy-shaping. 
In this study, the authors show that users who provide policy-shaping advice to the agents give more accurate advice and are willing to help the agent for longer, giving them more advice per episode.

%Figura PolicyShaping

\section{Proposed Architecture}

\begin{figure*}[t]
  \centering
    \includegraphics[width=1\linewidth]{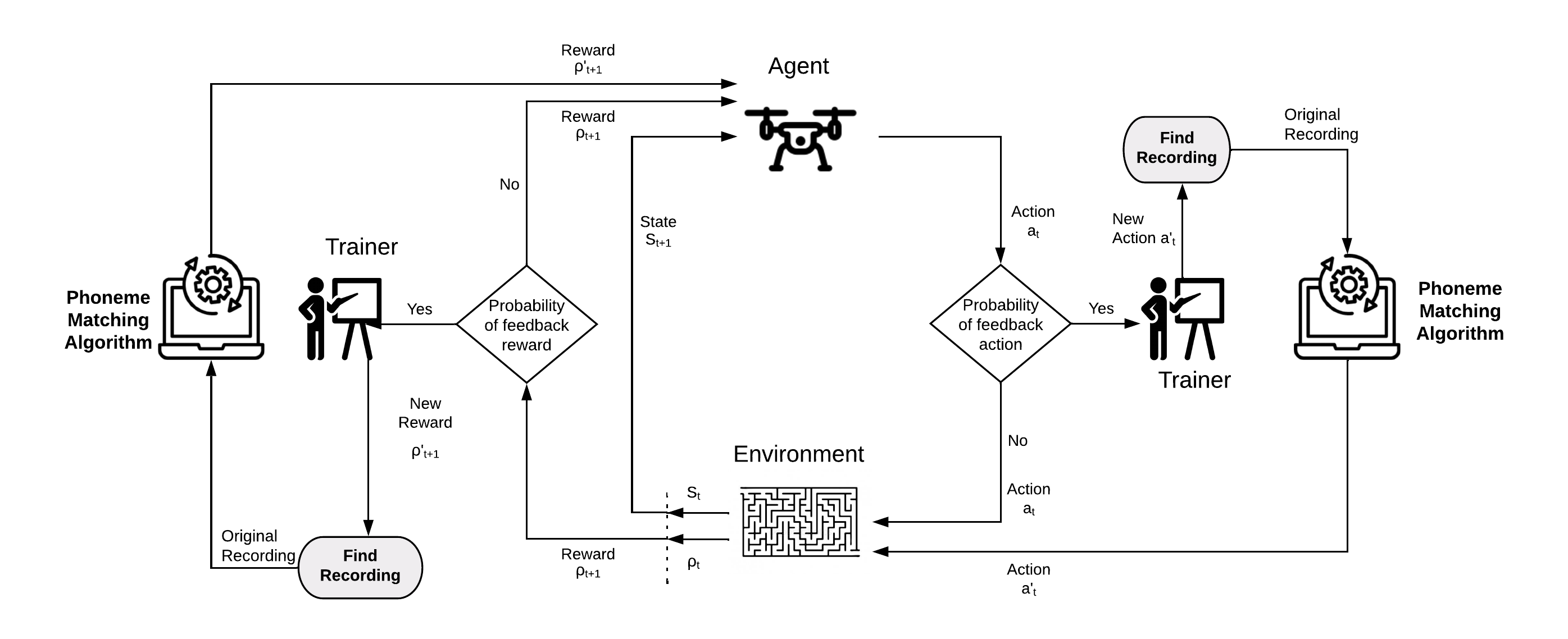}
   \caption{Proposed architecture for the IRL agent. The agent receives advice from two sources of variation. First, policy-shaping (right), where the trainer with probability $\mathcal{L}_a$ modifies the action selected by the agent. Second, reward-shaping (left), where the trainer with probability $\mathcal{L}_\rho$ modifies the reward received from the environment.}
   \label{Fig:IRL2}
\end{figure*}

In the literature, interactive reinforcement learning contemplates 
an external source of variation, that is, a trainer that is external to the environment and provides feedback to the agent during training~\cite{millan2021robust}. 
The source of variation can fall under the reward-shaping approach, which modifies the reward, or the policy-shaping approach, which modifies the decision of the agent.

In the proposed architecture, the agent receives feedback from two sources of variation, one that falls under reward-shaping and another that corresponds to policy-shaping. 
The agent and the environment interact at each iteration $t$. 
At each step, the agent receives a representation of the state of the environment, $s_t$, and selects an action, $a_t$, that is available in the current state.

Fig.~\ref{Fig:IRL2} presents a diagram of the proposed architecture, where the agent interacts with the environment while being advised by an external trainer under both approaches.
Once the action is selected, policy-shaping is performed. 
Thus, with a probability of feedback action $\mathcal{L}_a$ the external trainer selects a new action $a_t^\prime$, which modifies the state. 
As a result of performing the action, the agent receives a reward value of $\rho_{t+1}$ and achieves a new state, $s_{t+1}$.
Once the agent receives the reward, the reward-shaping takes place. 
With a probability of feedback reward $\mathcal{L}_{\rho}$, the agent receives a new reward $\rho_t^\prime$ from the external trainer.

On the other hand, the phoneme-matching algorithm presented in~\cite{contreras2020unmanned} was used to process the advice given by the external trainer. 
The advice is given via an audio selected from a set of recordings. 
Google Cloud Speech (GCS) in combination with a domain-based language was used to transform the signal into text. 
The audio transmissions are received and sent to the cloud-based GCS via the Web Speech API, from which a sentence recognized as a hypothesis is obtained. 
The hypothesis and the domain-based dictionary are then compared using the Levenshtein distance and the instruction with the minimum distance is selected. 
Once the voice command is converted into text, the signal is processed and classified as a reward (in the case of reward-shaping) or an instruction for the UAV (in the case of policy-shaping).

\begin{figure*}[t]
  \centering
  \subfigure[Experimental scenario without obstacles.] 
  {
  \includegraphics[width=0.4\textwidth]{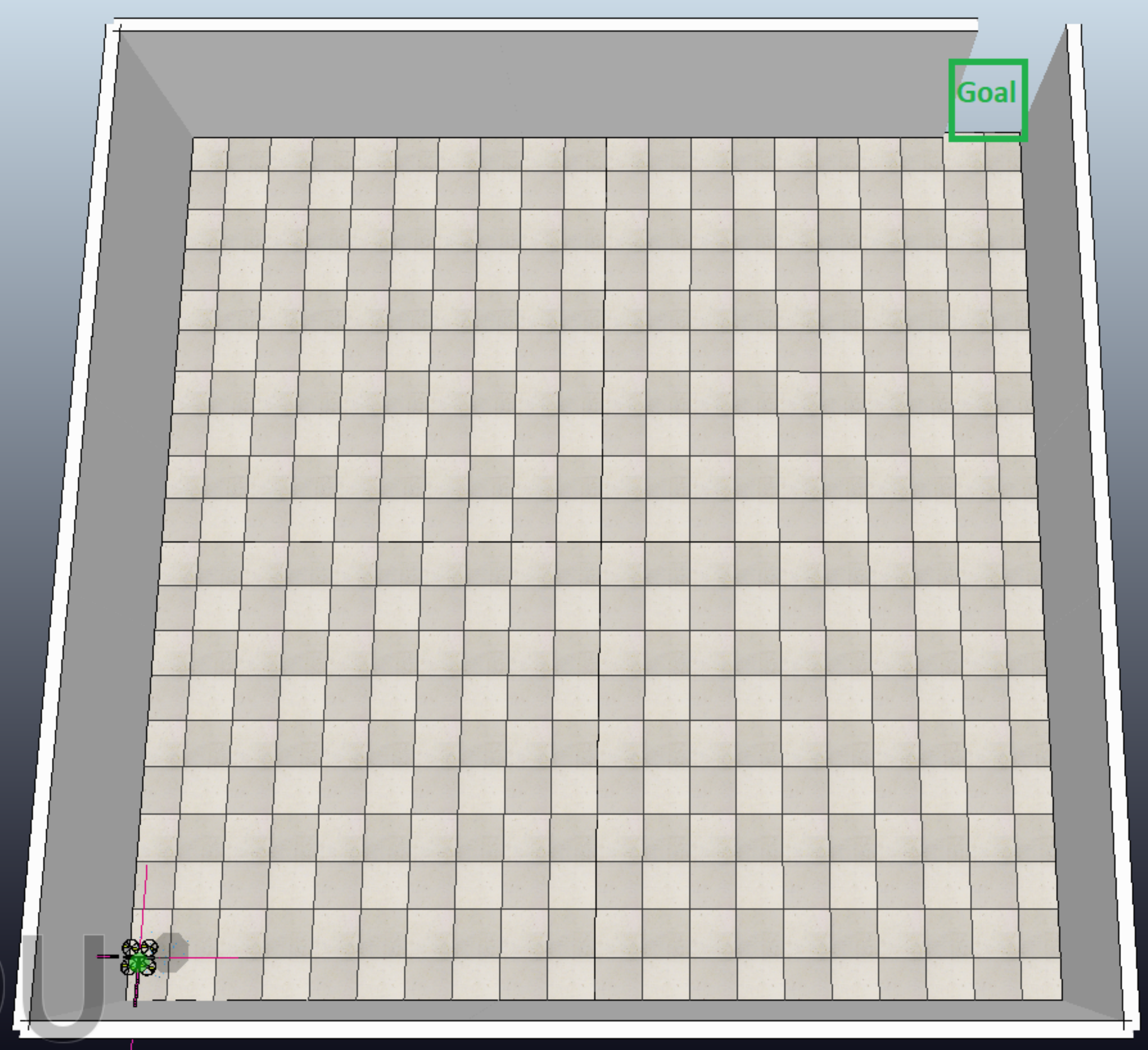} 
  \label{Fig:grilla-uno}
  }\hfill%space{0.5cm}
  \subfigure[Experimental scenario with obstacles.] {\includegraphics[width=0.4\textwidth]{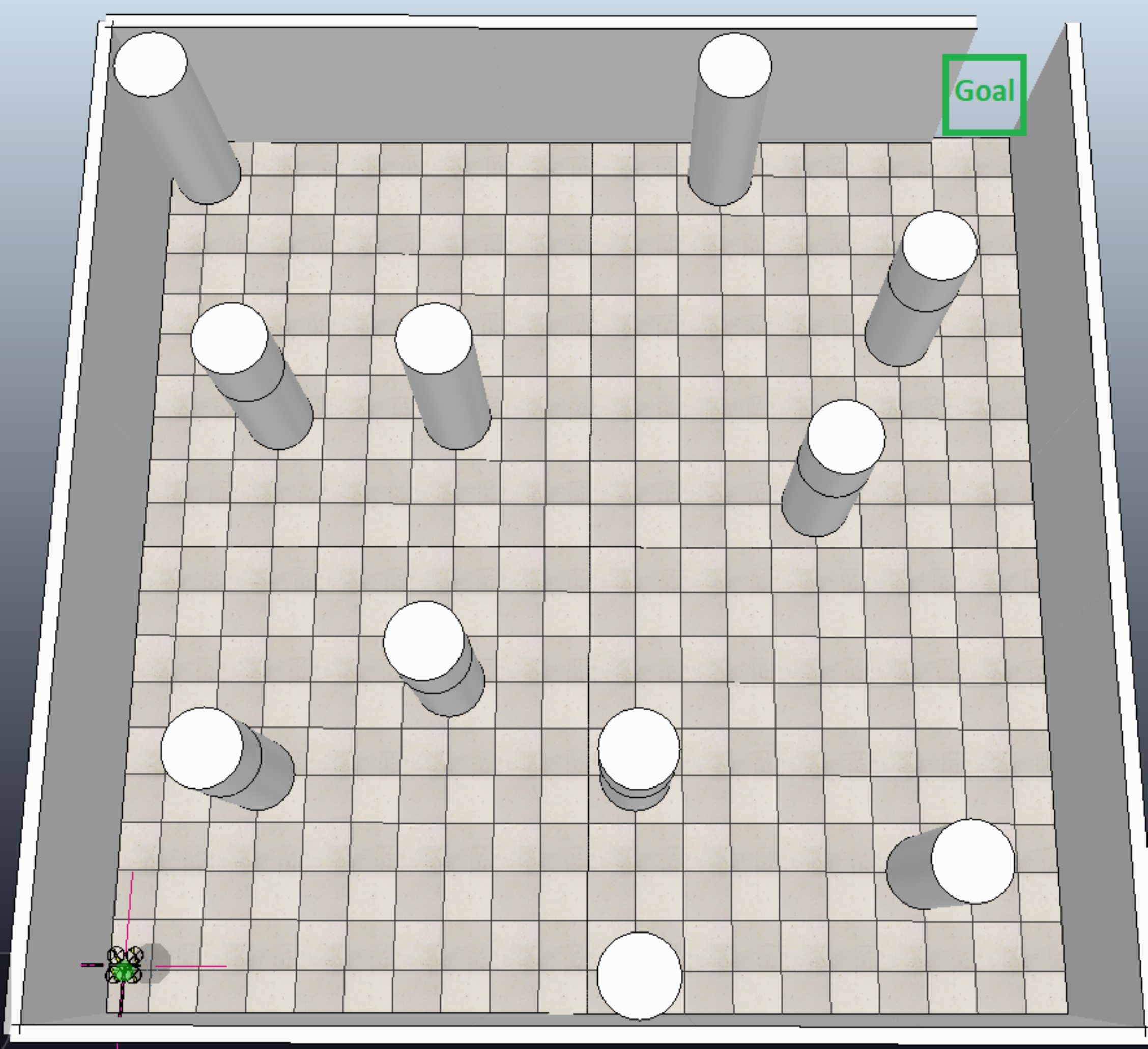} 
  \label{Fig:grilla-dos}
  }
  \caption{CoppeliaSim simulated environments. The UAV is located in the lower left corner and the target is found in the upper right corner. The task consists of the UAV navigating through the environment to the target while avoiding collisions with obstacles.}
  \label{Fig:simulacion-grilla}
\end{figure*}

The full implementation of the Q-learning algorithm with two sources of variation is shown in Algorithm~\ref{advice_algorithm}. 
The underlined lines show the policy-shaping advice (lines 6–8) and reward-shaping advice (lines 11–13). 
Unlike other implementations of reward-shaping, the reward is completely modified from the original, and is not increased (or decreased) by one value. 
However, the same behavior is maintained, giving it a more negative value when it fails and a more positive value when it succeeds.

\begin{algorithm}[t]
\caption{Q-learning with two sources of variation}
\label{advice_algorithm}
\begin{algorithmic}[1]
\REQUIRE $\gamma$, $\alpha$, $\epsilon$, $\mathcal{L}_{\rho}$, $\mathcal{L}_a$
\STATE Initialize $Q$ arbitrarily
\FOR{each episode}
	\STATE initialize $s_t$
		\REPEAT
			\STATE (Non advice) With probability $\epsilon$ select a random action $a_t$ %\\ \hspace{1.7cm} 
			otherwise select $a_t = \max_{a}Q(s_{t+1}, a_t)$
			\IF{\underline{$rand(0,1)<\mathcal{L}_a$}}
				\STATE \underline{Get advice $a^\prime_t$ based on speech recognition}
				\STATE \underline{Change action $a_t \leftarrow a^\prime_t$}
			\ENDIF
			\STATE Taken action $a_t$, observe reward $\rho_{t+1}$ and next state $s_{t+1}$
			\IF{\underline{$rand(0,1)<\mathcal{L}_{\rho}$}}
				\STATE \underline{Get advice $\rho^\prime_{t+1}$ based on speech recognition}
				\STATE \underline{Change reward $\rho_{t+1} \leftarrow \rho^\prime_{t+1}$}
			\ENDIF
			\STATE Update $Q(s_t, a_t) \leftarrow$ $Q(s_t, a_t) +$ \\ 
			\hspace{1.2cm} $\alpha \left[ \rho_{t+1} + \gamma \max_aQ(s_{t+1}, a_t) - Q(s_t, a_t) \right]$.
			\STATE $s_t \leftarrow s_{t+1}$
		\UNTIL{$s_t$ is terminal}
\ENDFOR
\end{algorithmic}
\end{algorithm}

\section{Experimental Environment}

To develop this project, we used CoppeliaSim~\cite{rohmer13vrep}, an open-source simulation software that provides a free educational license and can be used on several operating systems, such as Linux, Windows and iOS, to simulate different types of robots in realistic environments. 
It also has an ample range of API libraries to communicate with the simulator using different programming languages~\cite{ayala2020comparison}.

\subsection{Simulated scenarios}

Two scenarios were built for this project in order to test the proposed methodology. 
The two scenarios in the simulator can be seen in Fig.~\ref{Fig:simulacion-grilla}.

\begin{itemize}
    \item The first scenario (Fig.~\ref{Fig:grilla-uno}) is a $10 \times 10$ meter grid. There are no obstacles in the environment and therefore the drone can travel along free trajectories.
    \item The second scenario (Fig.~\ref{Fig:grilla-dos}) is a $10 \times 10$ meter grid that contains $11$ pillars of a diameter of $0.80$ m each, which are distributed throughout the environment, thus obstructing the path of the UAV along some trajectories.
\end{itemize}

At the start of the simulation, the UAV appears in the scenario at the bottom left (see Fig.~\ref{Fig:simulacion-grilla}) and will be able to move across the entire area, aiming to exit through the opening on the top right side (see the green rectangle in Fig.~\ref{Fig:simulacion-grilla}). 
The drone has four proximity sensors with a range of $1$ m each; they are located at the four cardinal points (“North”, “South”, “East”, and “West”).
In this way, if any object appears in front of any of these sensors, be it a wall or pillar, the drone will be able to detect a change of state. 
Thus, in the obstacle-free scenario, there are $400$ possible states ($10 \times 10 \times 4$), whereas in the scenario with obstacles there are $356$ possible states, since $11$ are occupied by the pillars, i.e., ($(10 \times 10 - 11) \times 4$).

Table~\ref{tab:instrucciones} shows the nine types of actions accepted by the UAV. 
Each action implies $1$ m of travel distance by the UAV; that is, if action $1$ or $4$ is selected (“Up” or “Go left”, respectively), the UAV will increase its altitude or move to the left by 1 m, respectively. 
The third column of Table~\ref{tab:rewards-des} shows the reward the agent receives. 
There are two negative values, corresponding respectively to a collision with an object ($\rho = -20$) or moving further away from the target ($\rho = -1$), and two positive values, which correspond to approaching the target ($\rho = 1.5$) and reaching the target ($\rho = 1000$).

\begin{table}[t]
    \caption{Commands to control the UAV and description of permitted actions.}
    \centering
    \begin{tabular}{cll}
        \toprule
        \textbf{No.} & \textbf{Action classes} & \textbf{Description}\\
        \midrule
        1  & Up           & Increases the altitude of UAV \\
        2  & Down         & Decreases the altitude of UAV \\
        3  & Go right     & Moves UAV to the right \\
        4  & Go left      & Moves UAV to the left \\
        5  & Go forward   & Moves UAV forward \\
        6  & Go back      & Moves UAV backwards \\
        7  & Turn right   & Turns UAV 90$^{\circ}$ clockwise \\
        8  & Turn left    & Turns UAV 90$^{\circ}$ counterclockwise \\
        9  & Stop         & Stops UAV movement \\ 
        \bottomrule
     \end{tabular}
     \label{tab:instrucciones}
\end{table}

The following constraints were imposed on the proposed scenarios to ensure a better performance of the simulation:

\begin{figure*}[t]
  \centering
  \subfigure[Average reward for the simulated scenario without obstacles.] 
  {
  \includegraphics[width=0.48\textwidth]{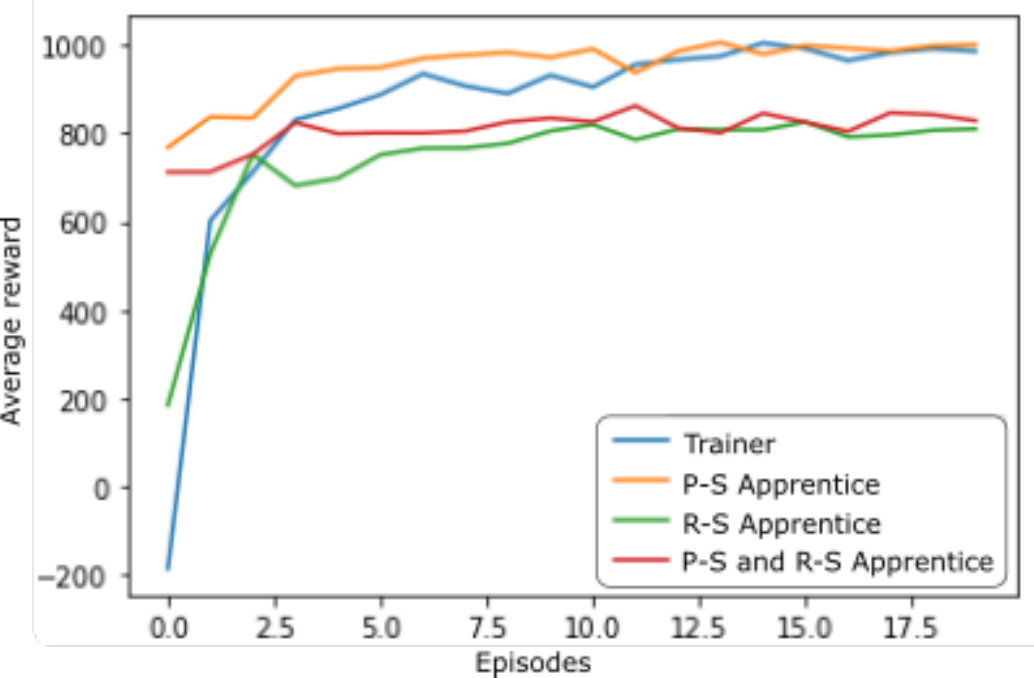} 
  \label{Fig:res-uno}
  }\hfill%space{0.5cm}
  \subfigure[Average reward for the simulated scenario with obstacles.] {\includegraphics[width=0.48\textwidth]{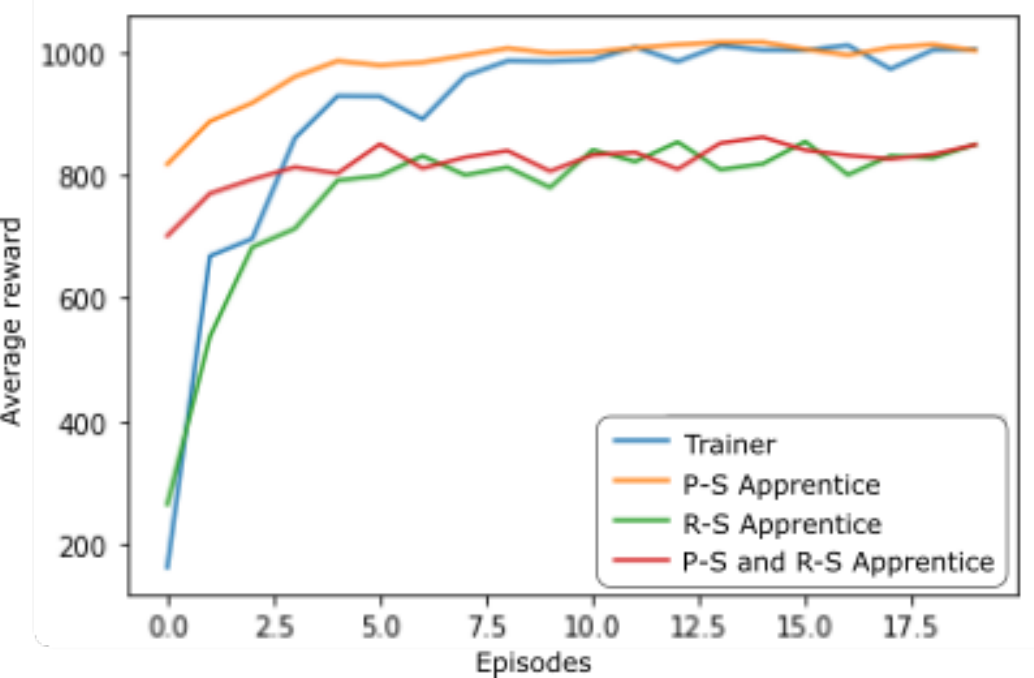} 
  \label{Fig:res-dos}
  }
  \caption{Average collected rewards for 20 RL agents using the different techniques in their respective simulated scenarios. In each plot, the blue line shows the performance of the autonomous agent used as trainer for IRL approaches.}
  \label{Fig:resultados-uno}
\end{figure*}

\begin{itemize}
    \item The maximum altitude of the UAV cannot exceed $2.50$ m; therefore, if the drone is at this altitude and is requested to climb, it will not perform the action.
    \item The minimum altitude of the UAV cannot be less than $0.50$ m; therefore, if the drone is at this altitude and is requested to descend, it will not perform the action.
    \item If the drone is requested to move in a direction and there is an object or wall that obstructs the passage of the UAV, the sensor in that direction will be activated and thus the drone will not execute the movement.
\end{itemize}

\subsection{External trainer and advice}

Nine classes that correspond to each action of the policy-shaping advice and four classes that correspond to the reward-shaping advice were created for the trainer. 
These are represented in a dictionary containing approximately $15$ instructions in Spanish and English. 
For each reward class, $15$ audios in Spanish and $15$ audios in English were recorded from a universe of five people. 
For policy-shaping, the trainer randomly selects an audio of the command corresponding to the advice (see Table~\ref{tab:instrucciones}, second column), which is processed by the phoneme matching algorithm to determine the action. 
For reward-shaping, the trainer randomly selects an audio from the corresponding class (see Table~\ref{tab:rewards-des}, fourth column), which is processed by the phoneme matching algorithm to determine which class it belongs to, awarding the corresponding reward.

\begin{table}[t]
    \caption{Values of the reward function (third column) and reward modification provided by the trainer (fourth column). The first column displays the evaluative commands supported by the phoneme matching algorithm.}
    \centering
    \begin{tabular}{llcc}
        \toprule
        \textbf{Class} & \textbf{Description } & \textbf{Reward} & \textbf{Reward-shaping} \\
        \midrule
        Very bad & Collision           & -20    & -10       \\
        Bad      & Further from target & -1     & -0.5      \\
        Well     & Closer to target    & 1.5    & 1         \\
        Perfect  & Target reached      & 1000   & 800       \\
        \bottomrule
     \end{tabular}
     \label{tab:rewards-des}
\end{table}

\subsection{Experimental Setup}

To evaluate the approach, four experiments were proposed and applied in both scenarios. The experiments are as follows:

\begin{itemize}
    \item $20$ autonomously trained RL agents
    \item $20$ IRL agents, with a probability of policy-shaping advice of $15\%$, where the trainer is one of the autonomously trained agents from the previous experiment.
    \item $20$ IRL agents, with a probability of reward-shaping advice of $15\%$, where the trainer is the same autonomously trained agent as in the previous experiment.
    \item $20$ IRL agents, with a probability of policy-shaping and reward-shaping advice of $15\%$, where the trainer is the same autonomous agent as in the previous experiments.
\end{itemize}

Each agent is trained for $20$ episodes using the Q-learning algorithm and the reward defined in Table~\ref{tab:rewards-des}.
The $\epsilon$-greedy action selection method is used with a value of $\epsilon = 0.1$. The values of discount factor $\gamma$ and learning rate $\alpha$ were set to $0.95$ and $0.1$, respectively. 
The experiments were conducted on a computer of the following characteristics: Intel Core i7-8750H processor, 8GB DDR4 2666MHz RAM, NVIDIA GeForce GTX 1050Ti with 4GB GDDR5 and Windows 10 Home. 
A fiber optic internet connection was used for the Web Speech API with an upload/download speed of 300/100 Mbps.

\section{Experimental Results}

\begin{table*}[t]
    \caption{Average and standard deviation of the reward obtained by the agents for each of the scenarios and the four experiments. The values in bold are the highest average reward value and the lowest standard deviation.}
    \centering
    \begin{tabular}{llccc}
        \toprule
        \textbf{Scenario} & \textbf{Experiment} & \textbf{Average Reward} & \textbf{Standard Deviation} \\
        \midrule
        Without obstacles & Autonomous    &  854.7      &  264.6   \\
                      & Policy-shaping        &  \textbf{952.2}      &  64.8    \\
                      & Reward-shaping        &  738.6      &  147.1   \\
                      & Policy-shaping and Reward-shaping   &  808.2      &  \textbf{40.5}    \\
        \midrule
        With obstacles & Autonomous    &  902.4      &  200.0   \\
                      & Policy-shaping         &  \textbf{979.6}      &  50.4    \\
                      & Reward-shaping        &  765.5      &  139.2   \\
                      & Policy-shaping and Reward-shaping   &  819.1      &  \textbf{35.8}    \\
        \bottomrule
     \end{tabular}
     \label{tab:accuracy-dos}
\end{table*}

This section presents the main results obtained, analyzing our proposed methodology in the two experimental scenarios.
Fig.~\ref{Fig:resultados-uno} shows the average reward per episode of the RL agents, for each of the scenarios. 
With respect to the scenario without obstacles (Fig.~\ref{Fig:res-uno}), it can be seen that the agents that had the slowest learning curve were the autonomous agents (blue line). 
The average reward of the agents trained with the policy-shaping technique (orange line) is slightly higher than that of the autonomous agents. 
However, the curve remains constant after episode $12$ and shows values that are similar to those of the autonomous agents. 
The agents trained with policy-shaping and reward-shaping simultaneously (red line) obtained lower rewards than the agents mentioned above but higher rewards than those agents trained with the reward-shaping technique.
However, after episode $10$, they remain constant. 
Table~\ref{tab:accuracy-dos}  shows the average total reward for each experiment as well as the standard deviation of the reward for the $20$ agents. 
The highest average total reward is achieved by the agents trained using policy-shaping, as is consistent with the observations in Fig.~\ref{Fig:res-uno}. 
In terms of standard deviation, the lowest is achieved by the agents trained with policy-shaping and reward-shaping, with a value of $40.5$. 
The highest standard deviation is seen in the autonomous agents, with a value of $264.6$.

In the scenario with obstacles (Fig.~\ref{Fig:res-dos}), it can be observed that the agents trained with the policy-shaping technique obtained a higher reward than the autonomous agents in the first episodes but after episode $11$, they achieved similar reward values.
The agents trained with the policy-shaping and reward-shaping techniques simultaneously show a slightly constant behavior, having a similar behavior to the agents trained with reward-shaping after $5$ episodes. 
Finally, the agents trained with policy-shaping had a higher reward in all episodes, though the autonomous agents showed a similar behavior after episode 8. 
The highest average total reward is once again achieved by the agents trained with policy-shaping, consistent with the observations in Fig.~\ref{Fig:res-dos}. 
In terms of standard deviation, the agents trained with policy-shaping and reward-shaping simultaneously show less dispersion (see Table~\ref{tab:accuracy-dos}), as in the scenario without obstacles.

In general, the agents trained with reward-shaping are those with lower values, because their reward function is different from that of the autonomous agents and those implemented with policy-shaping. 
One of the most significant differences is that when the agent reaches the final target it is granted a reward of $800$, compared with the $1000$ received by the other agents. 
This difference has a negative influence on the behavior of the learning curves of the reward-shaping agents.

\begin{figure}[t]
  \centering
    \includegraphics[width=0.5\textwidth]{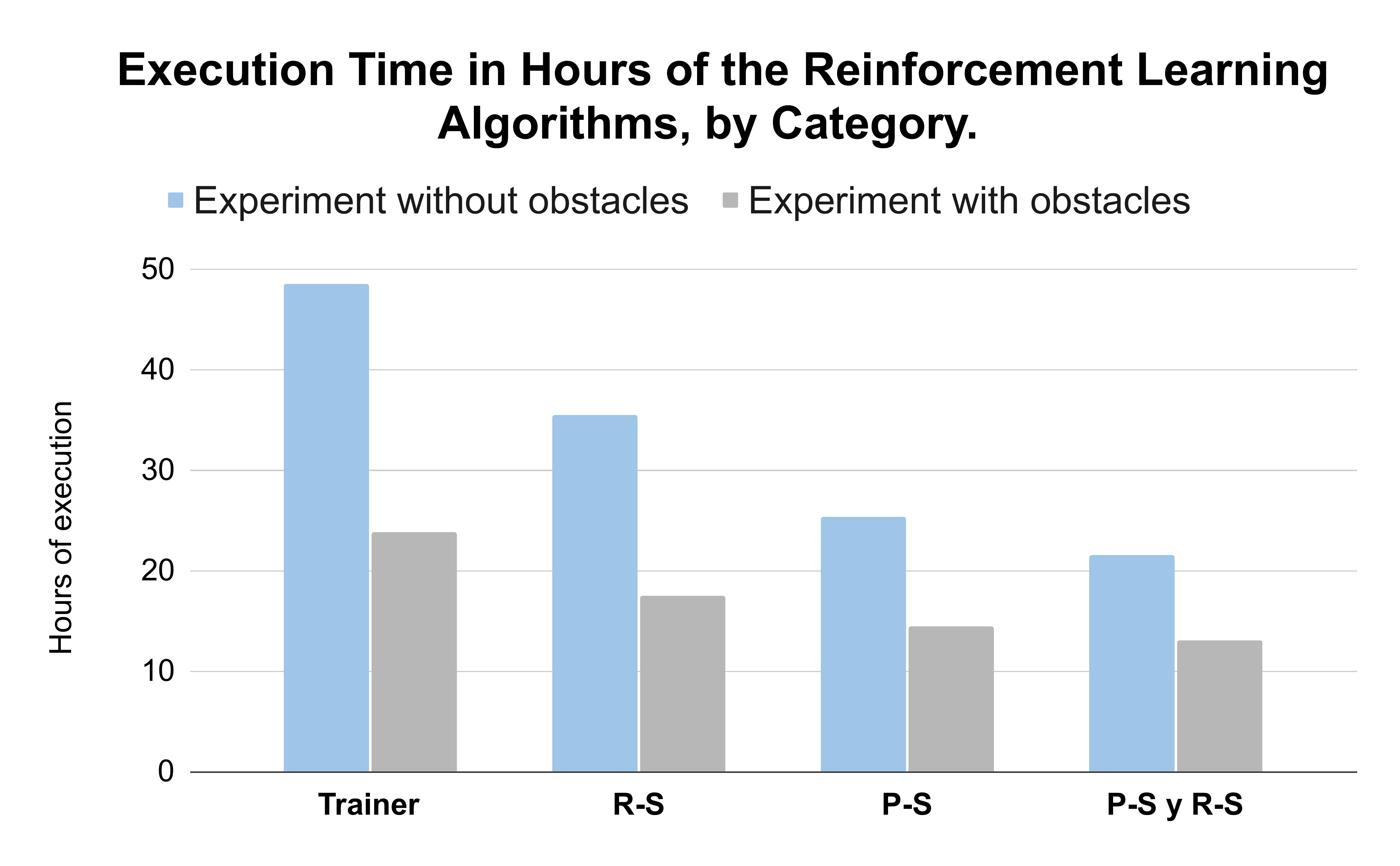}
   \caption{Execution times for each reinforcement learning algorithm, unit of measure is hours.}
   \label{Fig:T_ejec}
\end{figure}

Fig.~\ref{Fig:T_ejec} shows the execution times of each experiment in the two scenarios. The autonomous agents took longer to train in both scenarios, with execution times of $48.5$ and $23.8$ hours, respectively. 
The agents trained with policy-shaping and reward-shaping simultaneously had execution times of $21.6$ and $13.1$ hours for each scenario, respectively.
These values are $55.52\%$ and $44.96\%$ shorter, respectively, than the execution times of the autonomous agents. 
In general, the agents took less time in the scenario with obstacles, which is to be expected since the state space is smaller and can be explored by an agent in a shorter amount of time.

\section{Conclusions}

This paper presents an IRL approach to teach reinforcement learning agents using the reward-shaping and policy-shaping methods. 
The proposed approach was implemented in the control of a UAV in a simulated environment. 
The influence of the reward-shaping and policy-shaping methods on the performance of an agent, applied both individually and simultaneously, was evaluated during the training, and the execution times of each experiment were also compared.

In terms of average reward, in the experiment without obstacles, the agent trained with policy-shaping obtained the highest values in all episodes. 
However, after approximately $11$ episodes, the autonomous agent obtains similar values. 
On the other hand, the agent trained with reward-shaping and policy-shaping simultaneously obtains higher reward values than the autonomous agent in the first episodes, but does not maintain a rising trend, unlike the autonomous agent. 
With regard to the experiment with obstacles, similar behavior to the previous experiment is observed. 
In this experiment, there is a notable difference between the average reward of the autonomous agents and the policy-shaping agents compared with those of the reward-shaping agents and the agents trained with reward-shaping and policy-shaping simultaneously. 
The average reward curves of the policy-shaping agents show a rising trend in the first episodes, and those of the reward-shaping agents remain constant in the final episodes. 
The behavior suggests that reward-shaping advice does not benefit learning in the first episodes, though it does help to maintain good performance in the final episodes. 
By contrast, policy-shaping advice benefits learning in the first episodes. 
In both scenarios, on average, the agents trained with policy-shaping obtained higher rewards. 
However, the agents trained with policy-shaping and reward-shaping simultaneously showed lower dispersion, with lower standard deviation values. 
Given that policy-shaping provides information about the actions and not about their performance, the dispersion will be better for agents trained under this method. However, the reward-shaping method helps to maintain low dispersion in the last episodes. 
In terms of execution times, the agents trained with policy-shaping and reward-shaping simultaneously obtain shorter times, specifically, of about $17.35$ hours. 
Execution time for the policy-shaping method is lower than for reward-shaping, a behavior that was observed in the average reward.

In terms of future studies, we intend to implement this proposed approach in an experimental scenario of greater complexity and with environmental elements that are more closely related to a domestic environment. 
We are also working to implement the approach in a real environment, where the agent learns to control the UAV using deep learning techniques. 
We also intend to study the performance of the approach under other parameter settings. 
Furthermore, we aim to study the trade-off between policy-shaping and reward-shaping, in such a way that the agent receives advice from both sources of variation but at different stages of learning.

\section*{Acknowledgment}

This research was partially financed by Universidad Central de Chile under the research project CIP2020013, the Coordenação de Aperfeiçoamento de Pessoal de Nível Superior—Brasil (CAPES)—Finance Code 001, Fundação de Amparo a Ciência e Tecnologia do Estado de Pernambuco (FACEPE), and Conselho Nacional de Desenvolvimento Científico e Tecnológico (CNPq)—Brazilian research agencies.

\bibliographystyle{IEEEtran}
\balance
\bibliography{references}

\end{document}